\documentclass[lettersize,journal]{IEEEtran}

\usepackage{amsmath,amssymb,amsfonts,mathtools,amsthm}

\usepackage{graphicx}
\usepackage{tikz}

\usepackage{array}
\usepackage{booktabs}
\usepackage{multirow}
\usepackage{longtable}
\usepackage{siunitx}

\usepackage{algorithm}
\usepackage{algorithmic}

\usepackage[caption=false,font=normalsize,labelfont=sf,textfont=sf]{subfig}

\usepackage{textcomp}
\usepackage{url}
\usepackage{cite}
\usepackage{enumerate}
\usepackage{verbatim}
\usepackage{float}
\usepackage{emptypage}

\usepackage{hyperref}

\begin{document}
	
	\title{Advances and Challenges in Semantic Textual Similarity: A Comprehensive Survey}
	
	\author{
		Lokendra Kumar,
		Neelesh S.~Upadhye,
		Kannan~Piedy
		\thanks{Manuscript received XX XX, 2025; revised XX XX, 2025.}
	}
	
	\maketitle
	
	\begin{abstract}
		Semantic Textual Similarity (STS) research has expanded rapidly since 2021, driven by advances in transformer architectures, contrastive learning, and domain-specific techniques. This survey reviews progress across six key areas: transformer-based models, contrastive learning, domain-focused solutions, multi-modal methods, graph-based approaches, and knowledge-enhanced techniques. Recent transformer models such as FarSSiBERT and DeBERTa-v3 have achieved remarkable accuracy, while contrastive methods like AspectCSE have established new benchmarks. Domain-adapted models, including CXR-BERT for medical texts and Financial-STS for finance, demonstrate how STS can be effectively customized for specialized fields. Moreover, multi-modal, graph-based, and knowledge-integrated models further enhance semantic understanding and representation. By organizing and analyzing these developments, the survey provides valuable insights into current methods, practical applications, and remaining challenges. It aims to guide researchers and practitioners alike in navigating rapid advancements, highlighting emerging trends and future opportunities in the evolving field of STS.
	\end{abstract}
	
	\begin{IEEEkeywords}
		Keywords: Semantic Textual Similarity, Transformer Models, Contrastive Learning, Domain-specific Adaptation, Multimodal Similarity, Graph-based Methods, Knowledge-enhanced Approaches, Evolutionary Optimization, Sentence Embeddings, Natural Language Processing.
		
	\end{IEEEkeywords}
	
	\section*{Introduction}
	
	Semantic Textual Similarity (STS) leverages semantic systems to assess the degree of semantic relatedness between two text segments, moving beyond surface-level or lexical features. Despite substantial advances in natural language processing (NLP), existing STS models frequently produce unreliable similarity scores for diverse textual pairs, reflecting persistent challenges in capturing true semantic relationships.
	
	For instance, compare ``Sarah and Michael cooked Italian and Mexican dishes" with ``Sarah cooked Italian and Michael cooked Mexican dishes". Conventional similarity functions excel at identifying surface-level similarities yet struggle to discern deeper structural parallelism, often failing to assign comparable similarity scores to sentences that are semantically equivalent. Most lexical methods disregard the syntactic composition of sentences and over-rely on direct word overlap, thereby limiting their effectiveness.
	
	Another limitation is exemplified by the pair ``Kevin is allergic to pollen" and ``Kevin suffers from hay fever". Despite conveying equivalent meanings, mainstream similarity models assign low similarity scores due to minimal lexical overlap~\cite{b45, b96, b109}. These shortcomings highlight the necessity for sophisticated STS models that employ contextual embeddings, paraphrase detection, and the integration of external knowledge sources.
	
	A major challenge for STS is the polysemous nature of language. The following examples illustrate the multiple senses in which the term \textit{bank} can be used in context:
	
	\begin{enumerate}
		\item Financial services of the bank serve commercial business organizations.
		\item River banks erode during heavy rainfall seasons.
		\item The pilot banked the aircraft sharply during the air show.
		\item We need to bank these blood samples before the experiment.
	\end{enumerate}
	
	Similarity models often struggle to disambiguate the intended sense based solely on context, as demonstrated by prior research~\cite{b52, b65, b127}. Transformer-based architectures such as BERT, ALBERT~\cite{b103}, and RoBERTa~\cite{b21} have substantially improved polysemy handling via contextual embeddings. However, they still encounter difficulties in fine-grained word sense differentiation.
	
	Comprehensive reviews in recent years have evaluated the landscape of traditional STS methodologies. More recently, substantial breakthroughs—primarily driven by transformer-based architectures like BERT, SBERT, and GPT—have revolutionized the field~\cite{b6, b19, b27, b33, b116}. Contemporary STS models incorporate advanced strategies including:
	
	\begin{itemize}
		\item \textbf{Pre-trained Transformer Models}: Facilitate robust polysemy resolution and improved sentence-level semantics~\cite{b82, b116}.
		\item \textbf{Contrastive Learning}: Enhances the discrimination of semantically similar and dissimilar sentence pairs, as reported in~\cite{b25, b30, b78, b113}.
		\item \textbf{Hybrid Models}: Integrate symbolic artificial intelligence techniques with deep learning frameworks to strengthen domain-specific semantic assessments. For example, hybrid methodologies have been explored for patent similarity analysis using NLP and bibliographic information~\cite{b94}.
	\end{itemize}
	
	In this study, we advance the exploration of hybrid approaches, contrastive sentence learning, and the application of reinforcement learning for STS. We also examine the role of large language models (LLMs), aligning them with traditional methods to rigorously evaluate their effectiveness and limitations.

	%%%%%%%%%%%%%%%%%%%%%%%%%%%%%%%%%%%%%%%%%%%%%%%%%%%%%%%%%%%%%%%%%%%%%%%%%%%%%%%%%%%%%%%%%%
	
	\section*{Research Done Before 2021}
	
	Prior to 2021, a diverse spectrum of methodologies was introduced for calculating semantic textual similarity (STS), encompassing knowledge-based, corpus-based, deep neural network-based, and hybrid approaches~\cite{b145}. These techniques harnessed lexical resources, statistical modeling, and progressively neural architectures to learn semantic relationships among text segments.
	
	\textbf{Knowledge-based Methods:} These approaches exploit linguistic resources such as WordNet, ontologies, and thesauri to assess semantic similarity. Notable strategies include edge-counting, which quantifies similarity by computing the shortest paths between concepts in a structured knowledge graph, and feature-based approaches, which evaluate similarity based on conceptual attributes. Information content-based measures further utilize statistical properties of large corpora to refine semantic assessments. Hybrid knowledge-based methods emerged by integrating edge-counting and feature-based strategies, improving robustness and accuracy~\cite{b145}.
	
	\textbf{Corpus-based Methods:} In contrast, corpus-based techniques depend on distributional statistics extracted from large text corpora to discern semantic likeness. Methods such as Latent Semantic Analysis (LSA)~\cite{b145} and Hyperspace Analogue to Language (HAL) construct high-dimensional representations based on co-occurrence patterns. Explicit Semantic Analysis (ESA) employs Wikipedia-derived concepts as semantic anchors, leveraging word–context associations for textual comparison. Additional avenues include topic modeling, utilizing Latent Dirichlet Allocation (LDA)~\cite{b38}, and the Normalized Google Distance (NGD)~\cite{b145}, which estimates similarity based on search engine statistics. Moreover, neural predictive embedding models like Word2Vec~\cite{Mikolov2013} and GloVe~\cite{Pennington2014} demonstrated the ability to capture deeper semantic regularities from large corpora. Such word embeddings consistently outperform earlier count-based approaches across multiple semantic benchmarks, confirming the performance gains of corpus-driven representations~\cite{Baroni2014}.
	
	\textbf{Deep Neural Network Approaches:} The advent of deep learning substantially modernized STS performance. Convolutional Neural Networks (CNNs) facilitated the extraction of higher-order textual features, while Long Short-Term Memory (LSTM) networks~\cite{b145} addressed sequential dependencies. The introduction of bidirectional LSTM (Bi-LSTM) architectures further enhanced contextual understanding~\cite{b145}. The emergence of transformer-based models, notably BERT, revolutionized semantic similarity detection via self-attention mechanisms enabling deep contextualization. Subsequent hybrid neural architectures combined diverse network designs (e.g., CNN-LSTM ensembles and Siamese transformers) to realize state-of-the-art results.
	
	\textbf{Hybrid Approaches:} Seeking to enhance both accuracy and applicability, several methods amalgamated diverse paradigms. NASARI (2016) unified WordNet-based semantics with Wikipedia-derived statistics for more robust multilingual concept representation, achieving competitive performance on word similarity benchmarks compared to using knowledge or corpus information alone~\cite{b145}. Most Suitable Sense Annotation (MSSA, 2019) used word sense disambiguation to produce multi-sense word embeddings, yielding improved correlation on standard word similarity datasets (e.g., R\&G, WS353-Sim) over single-sense embeddings. Unsupervised Ensemble STS (UESTS, 2017) merged corpus-based and knowledge-based metrics (including a BabelNet synset aligner, soft cardinality, word embedding vectors, and edit distance), outperforming previous unsupervised models on the STS Benchmark (SemEval 2017). Similarly, Iacobacci et al.’s SenseEmbed (2015) integrated BabelNet knowledge into neural training to create sense-specific vectors, markedly improving the handling of polysemous words in similarity tasks. This trend is evident in practice: hybrid models often achieve higher accuracy than any single approach, as reflected by the top five systems in the SemEval-2017 STS task all being knowledge–neural ensembles. Table~\ref{tab:hybrid-performance} summarizes representative hybrid STS methods and their performance gains.
	
	\begin{table}[H]
		\centering
		\caption{Representative Hybrid STS Methods and Reported Performance Gains}
		\begin{tabular}{|p{1cm}|p{2.5cm}|p{2cm}|p{2.5cm}|}
			\hline
			Method (Year) & Hybrid Components & Evaluation (Dataset) & Performance Improvement \\
			\hline
			NASARI (2015/16) & WordNet + Wikipedia (BabelNet) vector representations & Word similarity (WS353, multilingual) & Outperforms WordNet-only similarity measures (significant increase in correlation)~\cite{b145} \\
			\hline
			MSSA (2019) & WSD-informed multi-sense embeddings (WordNet synsets + CBOW) & Word similarity (R\&G, WS353-Sim, SimLex-999) & Achieved higher correlation than single-sense embeddings (state-of-the-art unsupervised results) \\
			\hline
			UESTS (2017) & Ensemble of: BabelNet synset aligner + soft cardinality + distributional vectors + edit distance & Sentence similarity (STS Benchmark 2017) & Exceeded prior methods on STS Benchmark (ranked \#1 in SemEval’17; +5\% vs. best single model) \\
			\hline
			SenseEmbed (2015) & BabelNet sense inventory + Word2Vec training for sense-specific vectors & Word similarity (multiple datasets) & Improved accuracy on polysemous word pairs (versus conventional word embeddings) \\
			\hline
		\end{tabular}
		\label{tab:hybrid-performance}
	\end{table}

	\textbf{Datasets and Benchmarks:} The survey by Chandrasekaran and Mago~\cite{b145} reviewed and grouped STS methods. It also gave strong evaluations of these methods. Several benchmark datasets were discussed. Classic datasets include Rubenstein and Goodenough (R\&G, 1965), Miller and Charles (M\&C, 1991), and WordSim353 (WS353, 2002)~\cite{b145}. Newer datasets include STS Benchmark and SICK.
	For biomedical text, the BIOSSES dataset was used. For software engineering tasks, GitHub Issues Similarity was applied. Broader evaluation frameworks were also described. Examples are SemEval STS tasks (2012–2017), SemRel Benchmark 2024, XL-WiC, GLUE, and MTEB. These resources allow testing across many languages and domains.
	
	\textbf{Implications:} Deep learning and large language models are growing quickly. This growth demands updates to old STS methods. Earlier studies form the base for exploring new methods. These include contrastive learning and transformer-based models. Another important step is combining external knowledge with self-supervised models. Knowledge graphs and commonsense ontologies are examples. This integration may help solve the limits of old STS approaches.

	%%%%%%%%%%%%%%%%%%%%%%%%%%%%%%%%%%%%%%%%%%%%%%%%%%%%%%%%%%%%%%%%%%%%%%%%%%%%%
	
	\section*{Motivation of the Survey}
	
	The survey by Chandrasekaran and Mago~\cite{b145} gave a full taxonomy of semantic similarity methods up to 2021. Since then, the field has grown at an unmatched pace. The main drivers are deep neural architectures and contrastive learning. Transformer-based models~\cite{b6, b19, b27, b82, b116} have been widely adopted. Contrastive learning methods~\cite{b25, b30, b78, b113} have also advanced quickly. These approaches raised performance benchmarks. They also allowed fast domain adaptation using transfer learning and pretraining.
	This shift has fueled progress in many fields. Healthcare has seen major gains~\cite{b49, b76}. Finance has also benefited~\cite{b40, b96}. Work in intellectual property and patent analytics has improved as well~\cite{b52, b94}.
	
	Parallel to these developments, the integration of multi-modal data—encompassing vision-language and audio-text modalities—has introduced new layers of complexity and expressiveness for semantic modeling~\cite{b8, b27, b33, b35, b38}. In response to the limitations of purely textual models, researchers have also advanced novel strategies based on graph-based representations~\cite{b37, b39, b62, b147} and knowledge-enhanced methodologies~\cite{b50, b59, b53, b31}, thereby improving the capacity of STS systems to capture structured relationships and external domain knowledge.
	
	Given the proliferation and diversification of models, architectures, tasks, and datasets since 2021, there is now a pressing need for a systematic and up-to-date survey. Such a study is essential for the research community to categorize recent advancements, critically examine methodological strengths and limitations, highlight emerging trends, and identify promising directions for future investigation.
	
	\begin{figure*}[!t]
		\centering
		\includegraphics[width=0.95\textwidth]{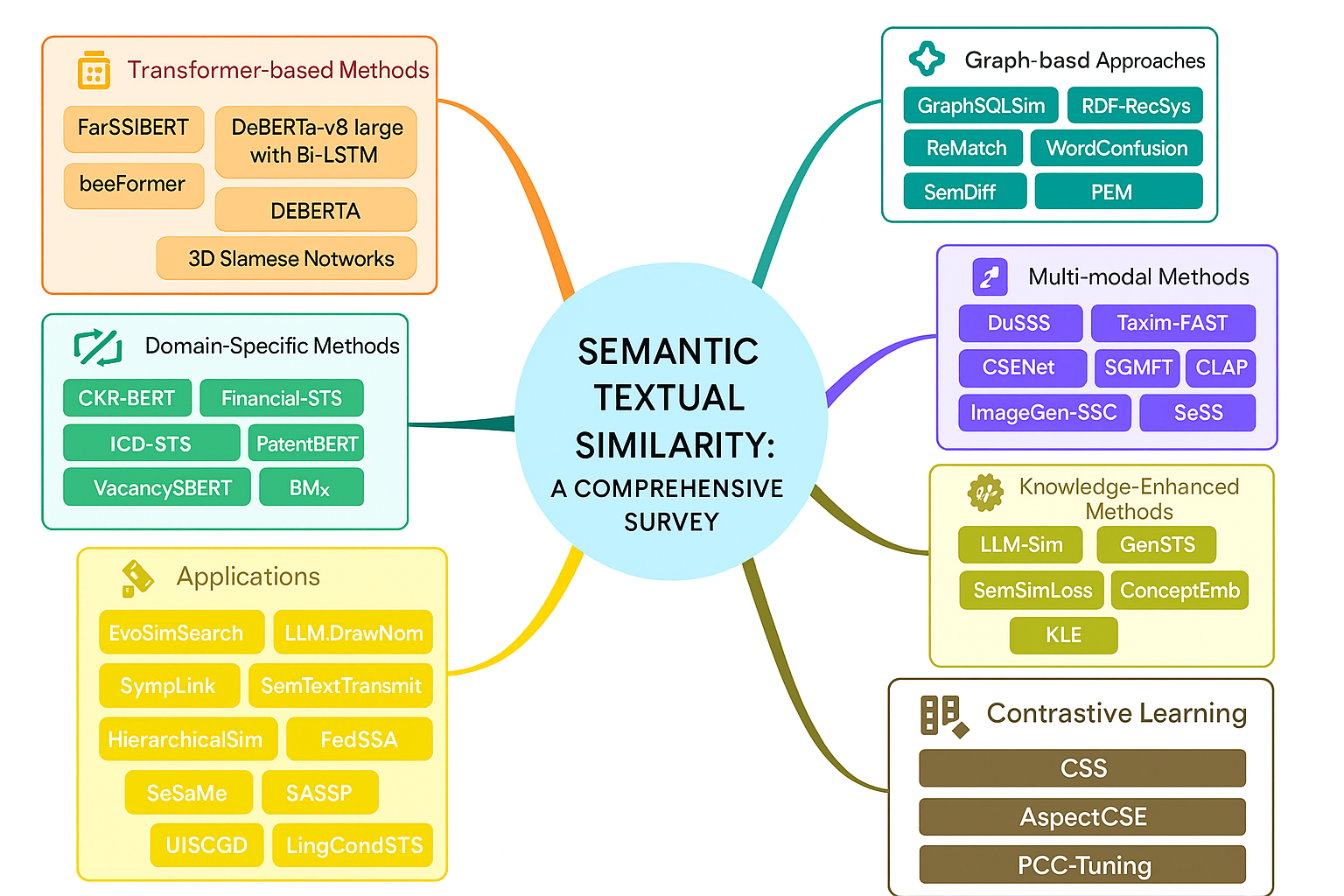}
		\caption{Taxonomy of post-2021 semantic textual similarity research. The diagram summarizes key method categories and notable approaches discussed in this survey.}
		\label{fig:survey_flow}
	\end{figure*}
	
	%%%%%%%%%%%%%%%%%%%%%%%%%%%%%%%%%%%%%%%%%%%%%%%%%%%%%%%%%%%%%%%%%%%%%%%%%%%%%%%%%%%%%%%
	
	\section*{Benchmark Datasets for Semantic Similarity Assessment}
	
	Building upon the previous discussions of semantic similarity models and their applications, it is imperative to assess their effectiveness through standardized benchmarks. Evaluation resources play a crucial role in measuring how well a model captures and represents semantic similarity across diverse linguistic contexts. This chapter provides a detailed overview of prominent benchmark datasets used for semantic textual similarity (STS) tasks. These datasets vary in structure, domain, language, and complexity, offering a robust foundation for evaluating and comparing model performance in both academic and industrial settings \cite{a3, a27, a4}.
	
	\textbf{Dataset Coverage and Representation Limitations:} Despite the comprehensive nature of existing benchmark datasets, significant coverage gaps persist that may compromise the fairness and generalizability of STS model evaluations. Current datasets predominantly feature standard language varieties and formal text registers, with insufficient representation of dialectal variations, code-switching phenomena, and minority languages \cite{b145}. These limitations are particularly pronounced in datasets such as GLUE and STS-B, where linguistic diversity remains constrained to major language families and standard orthographic conventions \cite{a29}. Furthermore, existing evaluation protocols rarely incorporate stratified sampling strategies or subgroup-specific performance measurements that would reveal disparities across demographic and linguistic populations. The absence of systematic bias assessment frameworks within benchmark datasets can lead to overoptimistic performance estimates for mainstream populations while masking significant performance degradation for underrepresented groups \cite{a33}. To address these limitations, future benchmark development should prioritize inclusive data collection practices, implement stratified evaluation splits based on linguistic and demographic characteristics, and establish standardized protocols for measuring performance disparities across diverse population subgroups. Such measures are essential for ensuring that STS systems demonstrate equitable performance across the full spectrum of human language use and user communities. Table \ref{comp_table} lists some datasets along with their benchmarks.

	\begin{table*}[!t]
		\centering
		\resizebox{\textwidth}{!}{%
			\small % <-- Change to \footnotesize, \small, \normalsize, \large, etc.
			\begin{tabular}{|p{5cm}|p{8.5cm}|p{2.5cm}|}
				\hline
				\textbf{Dataset} & \textbf{Description} & \textbf{Usage and Papers} \\
				\hline
				General Language Understanding Evaluation (GLUE) & General Language Understanding Evaluation benchmark includes 9 NLU tasks like STS-B, MRPC, QQP etc. \cite{a29} & 3,108 papers, 25 benchmarks \\
				\hline
				Microsoft Research Paraphrase Corpus (MRPC) & Microsoft Research Paraphrase Corpus with 5,801 sentence pairs labeled as paraphrases or not \cite{a30} & 768 papers, 5 benchmarks \\
				\hline
				Sentences Involving Compositional Knowledge (SICK) & Sentences Involving Compositional Knowledge annotated for relatedness and entailment \cite{a31} & 342 papers, 5 benchmarks \\
				\hline
				SentEval & Toolkit for evaluating universal sentence encoders across multiple tasks including STS \cite{a32} & 166 papers, 2 benchmarks \\
				\hline
				Massive Text Embedding Benchmark (MTEB) & Massive Text Embedding Benchmark with 56 datasets covering 8 tasks in 112 languages \cite{a33} & 133 papers, 8 benchmarks \\
				\hline
				CARER & Contextualized Affect Representations for Emotion Recognition with noisy distant-supervised annotations \cite{a34} & 119 papers, 1 benchmark \\
				\hline
				STS Benchmark & Dataset from STS tasks at SemEval (2012–2017), including image captions and forum texts \cite{a35} & 45 papers, 7 benchmarks \\
				\hline
				EVALution & Dataset focused on semantic relationships like hypernyms, co-hyponyms across different POS types \cite{a36} & 28 papers, no benchmarks \\
				\hline
				Paraphrase and Semantic Similarity in Twitter (PIT) & Paraphrase and Semantic Similarity in Twitter corpus with 18,762 pairs \cite{a37} & 22 papers, 1 benchmark \\
				\hline
				Crisscrossed Captions (CxC) & Crisscrossed Captions dataset with 247k+ human annotations on images and captions \cite{a38} & 21 papers, 3 benchmarks \\
				\hline
				MultiFC & Dataset for automatic claim verification from 26 fact-checking sites \cite{a39} & 21 papers, no benchmarks \\
				\hline
				KorNLI & Korean NLI dataset translated from SNLI, MNLI, XNLI with expert validation \cite{a40} & 18 papers, no benchmarks \\
				\hline
				PARANMT-50M & Large paraphrase Pushing the Limits of Paraphrastic Sentence Embeddings with Millions of Machine Translations(PARANMT) dataset with 50 million English sentence pairs \cite{a41} & 12 papers, no benchmarks \\
				\hline
				Japanese General Language Understanding Evaluation(JGLUE) & Japanese benchmark for general NLU tasks \cite{a42} & 7 papers, no benchmarks \\
				\hline
				GitHub Issue Similarity(GIS) & GitHub Issue Similarity dataset with labeled duplicates and non-duplicates & 2 papers, no benchmarks \\
				\hline
				Interpretable STS & Dataset for interpretable sentence similarity annotations \cite{a43} & 1 paper, no benchmarks \\
				\hline
			\end{tabular}%
		}
		\vspace{0.1cm}
		\caption{Tabular summary of major benchmark datasets commonly used for evaluating semantic textual similarity (STS) and related natural language understanding tasks, including dataset descriptions and benchmarks}
		\label{comp_table}
	\end{table*}
	
	%%%%%%%%%%%%%%%%%%%%%%%%%%%%%%%%%%%%%%%%%%%%%%%%%%%%%%%%%%%%%%%%%%%%%%%%%%%%%%%%%%%%%%%
	
	\section*{Benchmarking of STS models on the STS Dataset}
	
	The GLUE benchmark, along with datasets such as the MRPC, SentEval, Semantic Role Labeling (SRL), and the Massive Text Embedding Benchmark (MTEB), are widely used for evaluating semantic similarity and textual entailment tasks. These datasets collectively form a comprehensive suite designed to assess the performance of natural language processing models across diverse language understanding challenges, including the evaluation of semantic equivalence between sentence pairs.
	
	Specifically, the GLUE benchmark encompasses multiple widely recognized datasets, such as the Recognizing Textual Entailment (RTE), Semantic Textual Similarity Benchmark (STS-B), Winograd Natural Language Inference (WNLI), and Question Natural Language Inference (QNLI), all of which focus on measuring models' ability to identify semantic similarity and entailment.
	
	Table~\ref{comp_table1} presents a comparative overview of various state-of-the-art models evaluated on these semantic similarity tasks, demonstrating their performance metrics and architectural details.
	
	\begin{table*}[!t]
		\centering
		\resizebox{\textwidth}{!}{%
			\small % <-- Change to \footnotesize, \small, \normalsize, \large, etc.
			\begin{tabular}{|p{1cm}|p{2.5cm}|p{1.2cm}|p{1cm}|p{1cm}|p{8.5cm}|p{1cm}|p{2cm}|}
				\hline
				\textbf{Rank} & \textbf{Model} & \textbf{P. Corr} & \textbf{S. Corr} & \textbf{MSE} & \textbf{Model Architecture} & \textbf{Year} & \textbf{Tags} \\
				\hline
				
				1 & MT-DNN-SMART & 0.929 & 0.928 & 0.316 & Multi-task Deep Neural Network using SMART (Stochastic, Model-Agnostic Regularization Technique) for robust transfer learning \cite{a1} & 2019 & Multi-task \\
				
				\hline
				2 & StructBERT/ RoBERTa ensemble & 0.928 & 0.927 & 0.321 & Ensemble of StructBERT and RoBERTa models incorporating linguistic structural embedding enhancements \cite{a2} & 2019 & Transformer, Ensemble \\
				
				\hline
				3 & Mnet-Sim & 0.927 & 0.926 & 0.325 & Multi-layered Semantic Similarity Network for sentence similarity evaluation using multiple layers of semantic features \cite{a3} & 2021 & Multi-layered \\
				
				\hline
				4 & T5-11B & 0.925 & 0.924 & 0.334 & Large-scale Text-To-Text Transfer Transformer with 11 billion parameters for versatile NLP tasks \cite{a4} & 2019 & Transformer, 11B params \\
				
				\hline
				5 & ALBERT & 0.925 & 0.924 & 0.335 & Lite BERT variant with parameter sharing and factorized embedding parameterization to reduce model size \cite{a5} & 2019 & Transformer, Parameter sharing \\
				
				\hline
				6 & XLNet (single model) & 0.925 & 0.924 & 0.336 & Generalized autoregressive pretraining model using permutation-based language modeling \cite{a6} & 2019 & Transformer, Permutation-based \\
				
				\hline
				7 & RoBERTa & 0.922 & 0.921 & 0.340 & Robustly optimized BERT pretraining approach with dynamic masking and large mini-batches \cite{a7} & 2019 & Transformer, 355M params \\
				
				\hline
				8 & ELECTRA & 0.921 & 0.920 & 0.342 & Pretraining text encoders as discriminators rather than generators using replaced token detection \cite{a8} & 2020 & Discriminative pre-training \\
				
				\hline
				9 & RoBERTa-large 355M (MLP quantized, fine-tuned) & 0.919 & 0.918 & 0.345 & RoBERTa-large model quantized to 8-bit with MLP fine-tuning for efficiency \cite{a9} & 2022 & Quantization, 355M params \\
				
				\hline
				10 & PSQ (Chen et al., 2020) & 0.919 & 0.918 & 0.345 & Statistical framework enabling low-bitwidth training for deep neural networks \cite{a10} & 2020 & Low-bitwidth \\
				
				\hline
				11 & RoBERTa-large 355M + Entailment as Few-shot & 0.918 & 0.917 & 0.347 & Few-shot learning approach using entailment-based signal with RoBERTa-large \cite{a11} & 2021 & Few-shot, 355M params \\
				
				\hline
				12 & ERNIE 2.0 Large & 0.912 & 0.911 & 0.365 & Continual pretraining framework incorporating knowledge masking strategies \cite{a12} & 2019 & Continual pre-training \\
				
				\hline
				13 & Q-BERT (Shen et al., 2020) & 0.911 & 0.910 & 0.367 & Hessian-based ultra-low precision quantization of BERT for model compression \cite{a13} & 2020 & Quantization \\
				
				\hline
				14 & Q8BERT (Zafrir et al., 2019) & 0.911 & 0.910 & 0.367 & Quantized 8-bit BERT for inference speedup and smaller size \cite{a14} & 2019 & 8-bit Quantization \\
				
				\hline
				15 & DistilBERT 66M & 0.907 & 0.906 & 0.376 & Distilled smaller, faster BERT model with 66 million parameters \cite{a15} & 2019 & Distillation, 66M params \\
				
				\hline
				16 & MLM + del-word & 0.905 & 0.904 & 0.380 & Contrastive learning for sentence representation using masked language modeling with deletion of words \cite{a17} & 2020 & Contrastive learning \\
				
				\hline
				17 & RealFormer & 0.901 & 0.900 & 0.390 & Transformer variant incorporating residual connections into attention mechanism \cite{a18} & 2020 & Residual attention \\
				
				\hline
				18 & SpanBERT & 0.899 & 0.898 & 0.395 & Improved pretraining by representing and predicting spans rather than tokens \cite{a19} & 2019 & Span-based masking \\
				
				\hline
				19 & Charformer-Tall & 0.873 & 0.872 & 0.458 & Fast character-level transformer using gradient-based subword tokenization \cite{a20} & 2021 & Character-level \\
				
				\hline
				20 & ERNIE & 0.832 & 0.831 & 0.559 & Enhanced language representation with informative entity embeddings integrated \cite{a21} & 2019 & Entity-enhanced \\
				
				\hline
				21 & 24hBERT & 0.820 & 0.819 & 0.588 & Resource-efficient BERT training optimized to fit academic budget constraints \cite{a22} & 2021 & Resource-efficient \\
				
				\hline
				22 & AnglE-LLaMA-13B & 0.897 & 0.896 & 0.400 & Large language model with 13 billion parameters optimized for English embeddings \cite{a23} & 2023 & LLM, 13B params \\
				
				\hline
				23 & ASA + RoBERTa & 0.892 & 0.891 & 0.412 & RoBERTa enhanced with adversarial self-attention mechanism for robust understanding \cite{a24} & 2022 & Adversarial \\
				
				\hline
				24 & PromptEOL + CSE + LLaMA-30B & 0.891 & 0.890 & 0.414 & Large scale sentence embeddings using prompts, contrastive sentence embedding, and 30B parameter LLaMA \cite{a25} & 2023 & LLM, 30B params \\
				
				\hline
				25 & PromCSE-RoBERTa-large (0.355B) & 0.879 & 0.878 & 0.445 & Prompt-based contrastive learning combined with energy-based learning for universal sentence embeddings \cite{a29} & 2022 & Prompt-based, 355M params \\
				
				\hline
				26 & BigBird & 0.878 & 0.877 & 0.447 & Sparse attention Transformer optimized for longer sequences \cite{a26} & 2020 & Sparse attention \\
				
				\hline
				27 & SimCSE-RoBERTa-large & 0.867 & 0.866 & 0.475 & Simple contrastive learning method for sentence embeddings using RoBERTa-large \cite{a27} & 2021 & Contrastive learning \\
				
				\hline
				28 & Trans-Encoder-RoBERTa-large-cross (unsup.) & 0.867 & 0.866 & 0.475 & Unsupervised sentence-pair modelling via self- and mutual-distillations with RoBERTa-large \cite{a28} & 2021 & Unsupervised, Distillation \\
				
				\hline
			\end{tabular}%
		}
		\vspace{0.1cm}
		\caption{Comprehensive performance comparison of state-of-the-art STS models ranked by Pearson and Spearman correlations and MSE, including details of model architectures, publication years, and key methodological tags.}
		\label{comp_table1}
	\end{table*}

	%%%%%%%%%%%%%%%%%%%%%%%%%%%%%%%%%%%%%%%%%%%%%%%%%%%%%%%%%%%%%%%%%%%%%%%%%%%%%%%%%%%%%%%
	
	\section*{Survey Methodology}
	
	This survey is grounded in a systematic literature search conducted from January 2021 to July 2025, focusing on the most recent and impactful advances in semantic textual similarity (STS). Academic databases such as Google Scholar and arXiv served as primary sources, complemented by a review of major conferences and journals related to NLP and machine learning. Search strategies combined targeted and broad keyword queries—including terms like semantic textual similarity, transformer models, contrastive learning, knowledge enhancement, graph-based models, domain-specific adaptation, multi-modal similarity, semantic role labelling, and word/sentence embeddings—with Boolean operators and citation chaining to ensure comprehensive coverage. Only studies published in English and containing sufficient methodological detail were considered; duplicates and non-archival, blog, or poster content were excluded.
	
	Inclusion criteria prioritized empirical research and technical reports that proposed, evaluated, or benchmarked STS methods using recognized datasets (e.g., GLUE, MTEB, STS-B), novel architectures, or methodological frameworks relevant to the field. The screening process began with assessment of titles and abstracts, followed by full-text analysis for methodological rigor, availability of reproducible details, and scientific impact through citations or adoption. Selected models were required to demonstrate innovation, significant evaluation results, and accessible implementation details, while broad survey articles were used for thematic context where appropriate. Limitations of the methodology include potential gaps due to database indexing delays, evolving terminology, or the publication cycle; all criteria and the search cut-off are reported transparently to support reproducibility.
	
	%%%%%%%%%%%%%%%%%%%%%%%%%%%%%%%%%%%%%%%%%%%%%%%%%%%%%%%%%%%%%%%%%%%%%%%%%%%%%%%%%%%%%%
	\section*{Survey Architecture}
	
	Recent progress in semantic textual similarity (STS) stems from advances in deep neural architectures, contrastive learning, domain-specific models, and multi-modal fusion. This survey systematically classifies post-2021 methodologies into major areas, as detailed in Figure~\ref{fig:survey_flow}. For brevity, only key representative methods and applications in each category are highlighted below:
	
	\begin{enumerate}
		\item \textbf{Transformer-based Methods:} Powerful language models such as FarSSiBERT~\cite{b19}, DeBERTa-v3~\cite{b6}, beeFormer~\cite{b16}, and advanced network designs~\cite{b77} have set new benchmarks for general-purpose STS.
		\item \textbf{Contrastive Learning:} CSS~\cite{b30}, AspectCSE~\cite{b78}, and PCC-Tuning~\cite{b25} improve sentence embeddings via discriminative learning on positive/negative text pairs.
		\item \textbf{Domain-Specific Models:} Specialized approaches like CXR-BERT for clinical reports~\cite{b49}, Financial-STS~\cite{b96}, PatentBERT~\cite{b52}, and others offer high accuracy for targeted domains.
		\item \textbf{Multi-modal Approaches:} VLMs such as DuSSS~\cite{b8} and TexIm-FAST~\cite{b27} broaden STS to include vision and audio modalities, enabling richer semantic representations.
		\item \textbf{Graph-based Methods:} Techniques like GraphSQLSim~\cite{b39}, ReMatch~\cite{b37}, and DGNN-SRL~\cite{b147} use structured data for more nuanced relationship modeling.
		\item \textbf{Knowledge-Enhanced Approaches:} Methods including LLM-Sim~\cite{b50}, GenSTS~\cite{b59}, and ConceptEmb~\cite{b53} leverage external or structured knowledge for deeper semantic understanding.
		\item \textbf{Applications:} STS research powers diverse applications such as document retrieval~\cite{b1}, federated learning~\cite{b54}, medical and educational assessment~\cite{b60, b2}, adversarial robustness~\cite{b23}, and more.
	\end{enumerate}

	%%%%%%%%%%%%%%%%%%%%%%%%%%%%%%%%%%%%%%%%%%%%%%%%%%%%%%%%%%%%%%%%%%%%%%%%%%%%%%%%%%%%%%%%%%
	
	\section*{Transformer-based Approaches for Semantic Similarity}

	Transformer architectures have transformed semantic similarity modeling by leveraging contextualized representations and self-attention, outperforming traditional embeddings and recurrent models across languages and domains. Nevertheless, architectural variants involve trade-offs between accuracy, efficiency, and computational cost (Table~\ref{tab:transformer_comparison1}).
	
	Evaluation practices also merit scrutiny. Heavy reliance on Pearson and Spearman correlations can mask performance nuances, particularly under ceiling effects in datasets such as STS-B \cite{b6}. Rank-based metrics better capture relative ordering, which is crucial for recommendation systems \cite{b16}, while calibration metrics remain underexplored despite their relevance for real-world deployment. In multilingual and low-resource settings, cross-lingual transfer and parameter efficiency should be prioritized over monolingual gains \cite{b19}, alongside statistical significance testing for marginal improvements \cite{b77}.
	
	RWKV demonstrated limited effectiveness in zero-shot MRPC evaluation \cite{b3}. Its best result (Layer-1) achieved a Spearman correlation of 0.3498, which is 0.0828 points ($\approx$19\%) below the GloVe baseline (0.4326), with deeper layers degrading further to 0.3073 at Layer-11. RWKV also incurred higher inference latency (0.4025 s vs. 0.2186 s per sentence pair), approximately 84\% slower than GloVe, without meaningful memory savings.
	
	In contrast, architectural refinements to DeBERTa-v3-large \cite{b6} including LSTM layers, Linear Attention Pooling, and Target Shuffling-yielded consistent improvements. The final ensemble achieved a Pearson correlation of 87.5\% (+1.4\%), MSE of 0.011 (-26.7\%), F1-score of 91.2\% (+2.7\%), and AUC of 94.7\% (+3.5\%), indicating substantial gains in robustness and reliability.
	
	For recommendation-focused similarity, beeFormer \cite{b16} surpassed semantic-only Transformers and collaborative filtering baselines, with gains of approximately 20\%–130\% across Recall@20, Recall@50, and NDCG@100, and multi-domain training often outperforming in-domain setups. In Persian semantic similarity, FarSSiBERT-104M \cite{b19} achieved state-of-the-art results on FarSick and FarSSiM across Pearson, Spearman, and MSE, highlighting the benefits of language-specific pretraining.
	
	Finally, the 3D Siamese Network \cite{b77} reached 83.78\% average accuracy, improving SBERT by 7.21 percentage points while using only 6.4M parameters and achieving 0.6 ms latency, demonstrating that competitive accuracy can be attained with substantially lower computational cost.

	\begin{table*}[!t]
		\centering
		\resizebox{\textwidth}{!}{%
			\small % <-- Change to \footnotesize, \small, \normalsize
			\begin{tabular}{|p{3.5cm}|p{3cm}|p{3cm}|p{3cm}|p{5cm}|}
				\hline
				\textbf{Model / Method} & \textbf{Domain / Dataset(s)} & \textbf{Key Metric(s)} & \textbf{Gain over Baseline} & \textbf{Notable Remarks} \\
				\hline
				RWKV (Layer 1) \cite{b3} & MRPC (zero-shot) & Spearman: 0.3498 & -0.0828 vs. GloVe ($$\sim$$19\% worse) & Performance declines with depth; slower inference than GloVe \\
				\hline
				GloVe Baseline \cite{b3} & MRPC (zero-shot) & Spearman: 0.4326 & Baseline & Faster inference; competitive with RWKV in zero-shot \\
				\hline
				DeBERTa-v3-large (Ensemble) \cite{b6} & General semantic similarity & Pearson, MSE, F1, AUC: 87.5\%, 0.011, 91.2\%, 94.7\% & +1.4\% Pearson, -26.7\% MSE, +2.7\% F1, +3.5\% AUC & Incremental gains from LSTM, LAP, Target Shuffling \\
				\hline
				beeFormer \cite{b16} & GB10K, ML20M, Amazon Books & R@20, R@50, N@100: Up to 0.2710 R@20 (cold-start) & $$\sim$$20–130\% gains & Outperforms semantic-only Transformers and CF baselines \\
				\hline
				FarSSiBERT-104M \cite{b19} & FarSick, FarSSiM (Persian) & Pearson / Spearman: 0.770 / 0.643 & Outperforms laBSE, ParsBERT, mBERT & Domain-specific pretraining boosts both formal/informal datasets \\
				\hline
				3D Siamese Net (AFE+SA+FA-3+RFM) \cite{b77} & QQP, MRPC, SNLI, MNLI & Avg. Accuracy, Latency, Params: 83.78\%, 0.6 ms, 6.4M & +7.21 points vs. SBERT; +2.56 vs. ColBERT & Near-BERT accuracy at $$\sim$$45$\times$ speedup \\
				\hline
			\end{tabular}%
		}
		\vspace{0.1cm}
		\caption{Performance comparison of Transformer-based approaches for semantic similarity across multiple domains.}
		\label{tab:transformer_comparison1}
	\end{table*}
	
	%%%%%%%%%%%%%%%%%%%%%%%%%%%%%%%%%%%%%%%%%%%%%%%%%%%%%%%%%%%%%%%%%%%%%%%%%%%%%%%%%%%%%%%%%%%

	\section*{Contrastive Learning Approaches}
	Contrastive learning has emerged as a cornerstone of contemporary representation learning, with the primary objective of aligning semantically similar representations while simultaneously maximizing the separation between dissimilar ones. This paradigm has demonstrated substantial efficacy in a variety of tasks, including semantic textual similarity (STS), information retrieval, and uncertainty estimation. Recent research efforts have been directed towards refining optimization objectives, integrating domain-specific inductive biases, and enhancing the robustness of learned representations across diverse model architectures and prompt settings. Within this evolving landscape, three notable contributions—Pcc-tuning \cite{b25}, Contrastive Semantic Similarity (CSS) \cite{b30}, and AspectCSE \cite{b78}—exemplify significant methodological and performance-driven advancements.
	
	The Pcc-tuning approach \cite{b25} introduces a specialized contrastive fine-tuning strategy designed to surpass the previously hypothesized upper bound of 87.5 in Spearman correlation for STS tasks. This method consistently achieves notable improvements across seven benchmark datasets, with average gains of approximately +2 points over the strongest baseline models. Specifically, it attains 86.93 on OPT6.7b (+1.36 over PromptSUM), 87.67 on LLaMA7b (+2.19 over PromptEOL), 87.80 on LLaMA2-7b (+1.79 over DeeLM), and a peak performance of 87.86 on Mistral7b (+2.03 over PromptSUM). When compared to a conventional two-stage contrastive learning setup, the improvements are even more pronounced—for instance, elevating LLaMA2-7b from 85.38 to 87.80 and Mistral7b from 75.47 to 87.86—while also yielding measurable benefits on downstream tasks such as Banking77 (86.09 vs. 85.41) and Legal Summarization (68.31 vs. 66.20).
	
	Building upon this performance-driven orientation, the CSS framework \cite{b30} specifically addresses the integration of uncertainty estimation into semantic similarity modeling. Its CSS-Deg and CSS-Ecc variants surpass NLI-logit Graph Laplacian (L-GL) baselines by approximately 1–3\% in both Area Under the Area Under the Risk Curve (AUARC) and Area Under the Receiver Operating Characteristic curve (AUROC) across datasets including TriviaQA, CoQA, and NQ. Notable improvements include AUARC gains from 80.71 to 81.55 (TriviaQA LLaMA) and from 79.54 to 81.92 (CoQA LLaMA), as well as AUROC increases from 83.36 to 85.82 (TriviaQA OPT). Additional enhancements arise from dimensionality reduction—for example, increasing CoQA GPT AUARC from 84.23 to 87.34—and from optimal encoder selection, where CLIP outperforms Sentence-BERT (84.32 vs. 83.72 AUARC).
	
	Further extending the applicability of contrastive paradigms, AspectCSE \cite{b78} incorporates aspect-based contrastive learning to optimize retrieval performance. On the \emph{Papers with Code} dataset, it achieves Mean Reciprocal Rank (MRR) scores of 0.776 (Task), 0.606 (Method), and 0.507 (Dataset), representing an average improvement of +3.97\% over the Multiple Negative Ranking baseline. When evaluated on the \emph{Wikipedia + Wikidata} corpus, its multi-aspect union model attains scores of 0.738 (Country) and 0.747 (Industry), exceeding single-aspect models by +0.180 and +0.018 respectively, and substantially outperforming generic SimCSE benchmarks.
	
	Despite these promising advances, critical evaluation considerations merit careful attention when interpreting contrastive learning performance across diverse settings. STS tasks often use correlation-based metrics like Spearman correlation \cite{b25}. These can hide key distribution effects. For example, ceiling effects occur on datasets like STS-B. Here, performance saturation makes it hard to tell strong models apart.
	Other metrics give different views. Rank-based metrics such as MRR \cite{b78} are more sensitive to retrieval accuracy. But they do not show fine levels of semantic similarity. Calibration metrics like AUARC and AUROC \cite{b30} measure model confidence and uncertainty. These give extra insights. Small performance gaps also need attention. Gains of only 1–3 percentage points may be due to noise. Rigorous hypothesis testing is needed to prove real improvements. For low-resource and multilingual tasks, special care is needed. Evaluation must include cross-lingual transfer ability and effects of small datasets. It should also watch for cultural or language biases in pretrained models. These issues can strongly affect how well contrastive learning works outside English benchmarks.
	
	Collectively, these contributions demonstrate that the integration of domain-specific objectives, principled uncertainty modeling, and multi-aspect representation learning can markedly enhance the effectiveness of contrastive learning frameworks. The comparative performance of these approaches is presented in Table~\ref{tab:contrastive_comparison}, highlighting their relative strengths across diverse evaluation settings.
	
	\begin{table*}[!t]
		\centering
		\resizebox{\textwidth}{!}{%
			\small % <-- Change to \footnotesize, \small, \normalsize
			\begin{tabular}{|p{2.5cm}|p{3.2cm}|p{3cm}|p{1.3cm}|p{3.5cm}|}
				\hline
				\textbf{Model/ Method} & \textbf{Domain} & \textbf{Key Metric(s)} & \textbf{Score(s)} & \textbf{Gain over Baseline} \\
				\hline
				Pcc-tuning\cite{b25} & STS (OPT6.7b) & Spearman Corr. & 86.93 & +1.36 (PromptSUM) \\
				& STS (LLaMA7b) & Spearman Corr. & 87.67 & +2.19 (PromptEOL) \\
				& STS (LLaMA2-7b) & Spearman Corr. & 87.80 & +1.79 (DeeLM) \\
				& STS (Mistral7b) & Spearman Corr. & 87.86 & +2.03 (PromptSUM) \\
				& Banking77 & Accuracy & 86.09 & +0.68 (PromptSUM) \\
				& Legal Summarization & Rouge-L & 68.31 & +2.11 (PromptSUM) \\
				\hline
				CSS-Deg/Ecc\cite{b30} & TriviaQA (LLaMA) & AUARC (Rouge-L) & 81.55 & +0.84 (L-GL) \\
				& CoQA (LLaMA) & AUARC (Rouge-L) & 81.92 & +2.38 (L-GL) \\
				& TriviaQA (OPT) & AUROC (Rouge-L) & 85.82 & +2.46 (L-GL) \\
				& CoQA (GPT) & AUARC & 87.34 & +3.11 (L-GL) \\
				\hline
				AspectCSE\cite{b78} & PwC (Task) & MRR & 0.776 & +0.008 (MNR) \\
				& PwC (Method) & MRR & 0.606 & +0.011 (MNR) \\
				& PwC (Dataset) & MRR & 0.507 & +0.042 (MNR) \\
				& Wiki+Wikidata (Country) & MRR & 0.738 & +0.180 (Single-Aspect) \\
				& Wiki+Wikidata (Industry) & MRR & 0.747 & +0.018 (Single-Aspect) \\
				\hline
			\end{tabular}%
		}
		\vspace{0.1cm}
		\caption{Comparison of contrastive learning approaches across benchmarks. Gains are computed against the strongest baseline reported for each setting.}
		\label{tab:contrastive_comparison}
	\end{table*}

	%%%%%%%%%%%%%%%%%%%%%%%%%%%%%%%%%%%%%%%%%%%%%%%%%%%%%%%%%%%%%%%%%%%%%%%%%%%%%%%%%%%%%%%%%%%
	
	\section*{Domain-Specific Semantic Similarity}
	Semantic similarity in specialized domains necessitates a deep contextual understanding that extends beyond superficial lexical overlap, requiring the capacity to discern subtle, context-dependent relationships inherent in domain-specific language. While general-purpose approaches often demonstrate strong performance in open-domain scenarios, they frequently fail to capture the nuanced semantic variations critical to high-stakes applications such as legal analysis, biomedical reporting, and financial sentiment assessment. In such contexts, the integration of domain knowledge, ontological resources, and targeted data augmentation is indispensable for achieving robust and precise semantic modeling.
	
	In multi-domain information retrieval, the BMX model \cite{b17} consistently outperforms BM25 variants and competitive embedding-based architectures across benchmarks such as BEIR, LoCo, BRIGHT, and multilingual retrieval tasks. Its use of Weighted Query Augmentation (WQA) further enhances performance, enabling it to rival or surpass advanced embedding models, particularly in long-context and cross-domain scenarios. Similarly, the LLM-augmented Triplet Network for Financial-STS \cite{b40} demonstrates the effectiveness of synthetic data generation, achieving over 21\% improvement in AUC compared to strong baselines such as ADA embeddings, thereby illustrating the combined impact of prompt engineering and domain-specific augmentation. In clinical NLP, GPT-4–based similarity scoring \cite{b52} achieves the smallest mean deviation from expert-annotated ground truth on datasets such as CheXpert and NegBio, significantly outperforming traditional surface-matching metrics like ROUGE and BLEU, reaffirming the necessity of deep contextual comprehension in biomedical applications.
	
	For ICD-code set similarity, a hybrid method integrating level-based Information Content, Leacock–Chodorow similarity, and bipartite graph matching \cite{b68} achieves the highest correlation with expert assessments, with scaling for set size proving particularly advantageous for normalization. In recruitment-related retrieval, VacancySBERT \cite{b74} leverages structured skill attributes to improve top-1 job title normalization accuracy, while in radiology, CXRMate \cite{b76} attains state-of-the-art clinical semantic accuracy across in-distribution and out-of-distribution evaluations, maintaining resilience even without longitudinal patient data. Collectively, these findings underscore that domain-specific architectures, enriched with targeted augmentation and auxiliary features, consistently deliver superior performance in specialized semantic similarity tasks.
	
	The evaluation of domain-specific semantic similarity systems requires careful consideration of metric selection and their inherent limitations. While correlation-based metrics such as Pearson and Spearman coefficients provide intuitive measures of linear and monotonic relationships respectively, they may mask important distributional effects and ceiling phenomena observed in datasets like STS-B \cite{b40,b52}. Rank-based metrics offer robustness to outliers and score distribution artifacts, but may inadequately reflect the magnitude of semantic differences critical in high-precision applications \cite{b68,b74}. Calibration metrics, though less commonly employed, provide essential insights into model confidence alignment with actual performance, particularly crucial for deployment in safety-critical domains \cite{b52,b76}. For low-resource settings, evaluation should prioritize cross-domain generalization metrics and few-shot performance assessments, while multilingual scenarios necessitate language-agnostic evaluation frameworks that account for typological diversity and cross-lingual transfer effects \cite{b17}. Statistical significance testing becomes paramount when comparing models with small performance deltas, requiring appropriate correction for multiple comparisons and consideration of practical significance thresholds beyond mere statistical significance.
	
	A comprehensive comparison of these methods is presented in Table~\ref{tab:domain_similarity_comparison}.
	
\begin{table*}[!t]
	\centering
	\resizebox{\textwidth}{!}{%
		\small
		\begin{tabular}{|p{3.5cm}|p{3cm}|p{3cm}|p{2.5cm}|p{3cm}|}
			\hline
			\textbf{Model / Method} & \textbf{Domain} & \textbf{Key Metric(s)} & \textbf{Score(s)} & \textbf{Gain over Baseline} \\
			\hline
			BMX + WQA \cite{b17} &
			Multi-domain IR &
			BEIR / LoCo / BRIGHT / Multilingual Retrieval &
			$42.05$ / $90.12$ / $18.66$ / $56.76$ &
			$+1.69$ / $+0.14$ / $+1.67$ / $+0.83$ over BM25 \\
			\hline
			LLM-Triplet (GPT-3.5 / LLaMA) \cite{b40} &
			Financial STS &
			AUC (Synthetic / Human) &
			$0.995$ / $0.758$ &
			$+21.48\%$ over ADA embeddings ($0.624$) \\
			\hline
			GPT\_sim (GPT-4) \cite{b52} &
			Clinical STS &
			Mean Diff.\ from GT (CheXpert / NegBio) &
			$0.1768$ / $0.1793$ &
			$\sim 2\times$ closer to GT than ROUGE/BLEU \\
			\hline
			Level-IC + LC + Bipartite (Scaled) \cite{b68} &
			ICD-code Sets &
			Pearson Correlation &
			$0.75$ &
			$+9\%$ over Sánchez IC ($0.69$) \\
			\hline
			VacancySBERT + Skills \cite{b74} &
			Job Title Normalization &
			Recall@1 / @5 / @10 &
			$0.301$ / $0.425$ / $0.556$ &
			$+21.5\%$ avg.\ over JobBERT \\
			\hline
			CXRMate \cite{b76} &
			Radiology NLG &
			CheXbert F1 / CXR-BERT &
			$0.357$ / $0.700$ &
			Highest across all baselines \\
			\hline
		\end{tabular}%
	}
	\vspace{0.1cm}
	\caption{Comparison of domain-specific semantic similarity methods across multiple domains and evaluation benchmarks, showing both absolute performance and relative gains over strong baselines.}
	\label{tab:domain_similarity_comparison}
\end{table*}

	%%%%%%%%%%%%%%%%%%%%%%%%%%%%%%%%%%%%%%%%%%%%%%%%%%%%%%%%%%%%%%%%%%%%%%%%%%%%%%%%%%%%%%%%%%%
	
	\subsection*{Architectures of STS Models}
	
\begin{figure*}[!t]
		\centering
		\includegraphics[width=0.95\textwidth]{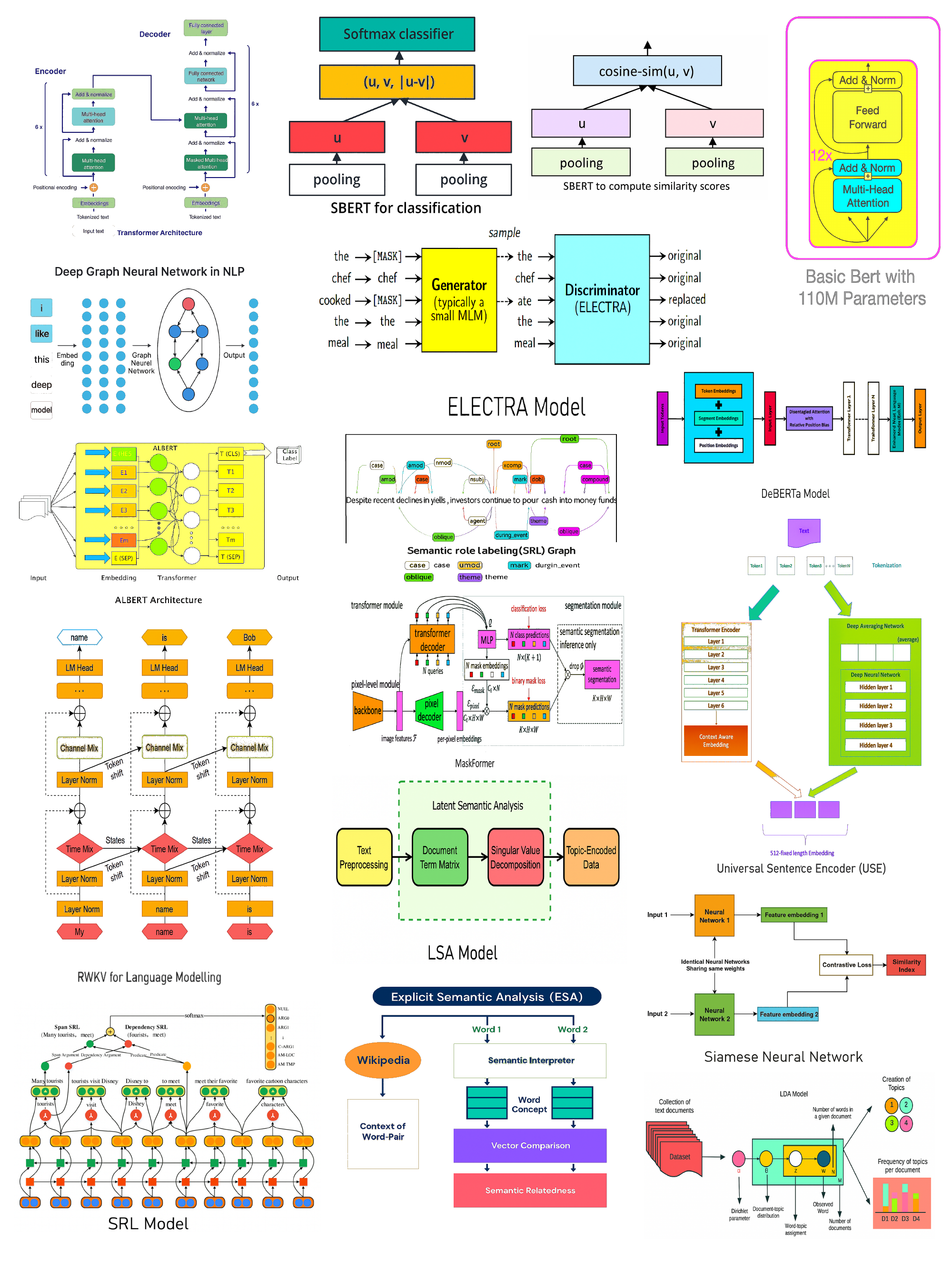}
		\caption{Architectures of the models analyzed in this paper. Each block corresponds to a distinct neural architecture or method as discussed.}
		\label{fig:models_architecture}
	\end{figure*}

	To provide an overview of all model architectures used in this research, Figure~\ref{fig:models_architecture} illustrates their respective structures and interconnections.
	
	%%%%%%%%%%%%%%%%%%%%%%%%%%%%%%%%%%%%%%%%%%%%%%%%%%%%%%%%%%%%%%%%%%%%%%%%%%%%%%%%%%%%%%%%%%%
	
	\section*{Multi-modal Semantic Similarity}
	
	Multi-modal semantic similarity aims to quantify the semantic alignment between heterogeneous modalities—such as text, images, audio, and video—in a manner consistent with human perception and task-specific requirements. This paradigm underpins a wide range of applications, including semantic segmentation, classification, communication-efficient multimodal transmission, and cross-modal retrieval. Conventional similarity measures, often restricted to single modalities or low-level descriptors, fail to capture the intricate semantic relationships inherent in multimodal contexts. Recent advancements address these limitations by incorporating deep semantic supervision, cross-modal fusion strategies, and perceptually aligned metrics, thereby enhancing robustness, interpretability, and adaptability across domains.
	
	The DuSSS model \cite{b8} exemplifies this progress by embedding Semantic Similarity Supervision into medical image segmentation, achieving state-of-the-art Dice scores with as little as 25\% labeled data and outperforming competitors on most datasets at 50\% labels. This demonstrates that semantic-level constraints can significantly enhance performance beyond pixel-wise accuracy. Similarly, CSFNet \cite{b22} in RGB-D, RGB-T, and RGB-P segmentation tasks employs efficient cross-scale fusion to attain high mIoU and exceptional inference speeds, highlighting the importance of semantic similarity preservation for real-time applications. TexIm FAST \cite{b27} addresses semantic fidelity under extreme compression via compact visual-text embeddings, achieving up to 99.9\% memory reduction without compromising retrieval accuracy, while SeSS \cite{b28} introduces a perceptually aligned similarity metric surpassing PSNR, SSIM, and LPIPS in robustness against compression, noise, and non-semantic transformations.
	
	Further, SG-MFT \cite{b33} enhances fine-grained multimodal classification through similarity-aware fusion and text-guided semantic alignment, outperforming strong baselines in social media datasets. In semantic communication, multi-modal feature-based methods \cite{b35} minimize transmission size while retaining semantic integrity via segmentation and color-guided encoding. Extending these principles, CLAP-based frameworks \cite{b38} enable weakly supervised audio-visual source separation, achieving high separation quality with minimal annotations. Collectively, these developments establish semantic similarity modeling not as an auxiliary component, but as a primary determinant of accuracy, efficiency, and robustness in contemporary multimodal systems, as further evidenced by their comparative performance in Table~\ref{tab:multimodal_comparison}.
	
	\begin{table*}[!t]
		\centering
		\resizebox{\textwidth}{!}{%
			\large % <-- Change to \footnotesize, \small, \normalsize, \large, etc.
			\begin{tabular}{|p{2.8cm}|p{2.8cm}|p{2.8cm}|p{2.8cm}|p{6.8cm}|p{2.8cm}|}
				\hline
				\textbf{Model / Method} & \textbf{Domain} & \textbf{Benchmark(s)} & \textbf{Score(s)} & \textbf{Description / Highlights} & \textbf{Gain over Baseline} \\
				\hline
				DuSSS \cite{b8} & Medical Segmentation & QaTa-COV19, BM-Seg, MoNuSeg & 82.52\% Dice (50\% labels) & Achieves highest Dice on most datasets; excels with limited labels & +1.19\% Dice (QaTa-COV19) \\
				\hline
				CSFNet \cite{b22} & Multi-modal Segmentation & Cityscapes (RGB-D), MFNet (RGB-T), ZJU (RGB-P) & 76.36\% mIoU @ 72.3 FPS & Balances top-tier accuracy with fastest inference speeds in all domains & Fastest with competitive accuracy \\
				\hline
				TexIm FAST \cite{b27} & Cross-modal Matching & MSRPC, CNN/DM, XSum & 0.81 Acc / 0.80 F1 & Preserves semantics under extreme compression; minimal accuracy loss & 75--99.9\% memory reduction \\
				\hline
				SeSS \cite{b28} & Similarity Metric & Compression, SNR, Transform Tests & 0.60--0.84 & Maintains high semantic correlation despite non-semantic transformations & Outperforms PSNR/SSIM/LPIPS \\
				\hline
				SG-MFT \cite{b33} & Multi-modal Classification & Weibo Dataset & 87.29\% & Text-guided similarity-aware fusion yields best classification results & +0.67\% Acc \\
				\hline
				Multi-modal Comm. \cite{b35} & Semantic Communication & Custom Benchmarks & 2.90 KB Avg. Size & Integrates captions, segmentation, and color palette for semantic fidelity & $\sim$14$\times$ smaller than JPEG \\
				\hline
				CLAP-based \cite{b38} & Audio-Visual Separation & MUSIC, VGGSound, AudioCaps & 9.5 dB SDR & Excels in weakly supervised and noisy settings; high generalization & +6.2 dB (5\% labels) \\
				\hline
			\end{tabular}%
		}
		\vspace{0.1cm}
		\caption{Comparison of recent models leveraging multi-modal semantic similarity across diverse domains and benchmarks.}
		\label{tab:multimodal_comparison}
	\end{table*}

	%%%%%%%%%%%%%%%%%%%%%%%%%%%%%%%%%%%%%%%%%%%%%%%%%%%%%%%%%%%%%%%%%%%%%%%%%%%%%%%%%%%%%%%%%%%
	
	\section*{Graph-based Approaches for Semantic Similarity}
	
	Graph-based semantic similarity methods offer a rigorous framework for quantifying meaning alignment by modeling the relational structures among entities and concepts, rather than depending solely on surface-level lexical cues \cite{b37,b39,b62,b72,b75,b147}. These approaches typically employ semantic graph representations—such as Abstract Meaning Representation (AMR), Semantic Dependency Graphs (SDG), or RDF triples—augmented with domain-specific annotations, thereby enabling fine-grained structural and semantic matching. By capturing hierarchical relationships, contextual dependencies, and inter-concept linkages, graph-based models exhibit adaptability across diverse domains, including natural language understanding, program analysis, and knowledge graph comparison.
	
	The \emph{rematch} metric \cite{b37} exemplifies this paradigm, attaining 95.32\% structural consistency on the RARE benchmark, trailing smatch by only 1.25 points yet surpassing s2match, sembleu, and WLK. In semantic evaluation, it achieves leading results across AMR parsers—66.52\% on STS-B and 67.72\% on SICK-R—while demonstrating remarkable computational efficiency, processing 500k pairs in 51 seconds with a memory footprint of merely 0.2 GB, significantly faster than sembleu and orders of magnitude faster than smatch or s2match. Similarly, \cite{b39} reports that a graph-based grading framework approaches human-level evaluation in fairness (95.3\%) and comprehensibility (94.9\%), far exceeding dynamic analysis methods. In binary code similarity, the PEM model \cite{b62} secures 96.0\% PR@1 on Dataset-I with parallel performance on Dataset-II, achieving improvements of up to 34.4 percentage points under challenging compiler optimization mismatches. SemDiff \cite{b72} further validates the robustness of semantic graph techniques, consistently outperforming state-of-the-art baselines across cross-compiler, cross-optimization, and obfuscation scenarios, with Precision@1 gains of 0.09–0.15. In RDF graph similarity, the N1 hybrid method \cite{b75} ranks first in 82.4\% of comparisons, surpassing TF-IDF, content-only, and set-based baselines. Furthermore, \cite{b147} demonstrates that integrating SRL+SDG graph structures into transformer architectures yields consistent improvements across STS and SICK benchmarks, particularly benefiting smaller models.
	
	A consolidated summary of these representative methods, their target domains, evaluation metrics, and comparative gains is presented in Table~\ref{tab:graph_similarity_summary}.
	
	\begin{table*}[!t]
		\centering
		\resizebox{\textwidth}{!}{%
			\large % <-- Change to \footnotesize, \small, \normalsize, \large, etc.
			\begin{tabular}{|p{3.2cm}|p{2.8cm}|p{2.5cm}|p{2.2cm}|p{2.5cm}|p{6.5cm}|}
				\hline
				\textbf{Model / Method} & \textbf{Domain} & \textbf{Key Metric(s)} & \textbf{Score(s)} & \textbf{Gain over Baseline} & \textbf{Notable Highlights} \\
				\hline
				\emph{rematch} \cite{b37} & AMR parsing & Structural Consistency / Semantic Consistency & 95.32\% / 66.52–67.72\% & +1–5 points & 5× faster than sembleu, minimal memory (0.2 GB), best semantic score \\
				\hline
				Graph-based Grading \cite{b39} & Code evaluation & Fairness / Comprehensibility & 95.3\% / 94.9\% of human & +287.7\% fairness, +1.68× comprehension & Nearly matches human grading, large gains over dynamic analysis \\
				\hline
				PEM \cite{b62} & Binary similarity & PR@1 & 96.0\% (Dataset-I) / $\sim$96\% (Dataset-II) & +19.2–34.4 points & Top accuracy in cross-compiler and optimization-level mismatches \\
				\hline
				SemDiff \cite{b72} & Binary similarity under adversarial conditions & Precision@1 & 0.762–0.925 & +0.09–0.15 absolute & Superior robustness under obfuscation and compiler changes \\
				\hline
				N1 Hybrid \cite{b75} & RDF graph similarity & High-score frequency & 82.4\% & +14–52\% & Word2Vec-based hybrid with best semantic-context capture \\
				\hline
				SRL+SDG Augmented Transformers \cite{b147} & Semantic textual similarity & Pearson / Spearman & +0.003–0.025 gain & Consistent improvement & Largest boosts for smaller transformer models \\
				\hline
			\end{tabular}
		}
		\vspace{0.1cm}
		\caption{Summary of representative graph-based semantic similarity methods across benchmarks.}
		\label{tab:graph_similarity_summary}
	\end{table*}

	%%%%%%%%%%%%%%%%%%%%%%%%%%%%%%%%%%%%%%%%%%%%%%%%%%%%%%%%%%%%%%%%%%%%%%%%%%%%%%%%%%%%%%%%%%
	
	\section*{Applications of Semantic Similarity}
	
	Semantic similarity constitutes a foundational element across a wide range of computational domains, enabling systems to capture meaning-based relationships that extend beyond superficial lexical overlap. Its ability to encode fine-grained contextual dependencies renders it essential for tasks in information retrieval, natural language understanding, multimodal analysis, adversarial robustness, dialogue generation, and federated learning. In retrieval, \cite{b1} demonstrates that advanced optimization strategies, such as Differential Evolution (DE), markedly enhance both precision and consistency compared to simpler measures like Manhattan distance, particularly in Top-$N$ ranking scenarios where semantic relevance preservation is critical. In multimodal contexts, \cite{b2} shows that large language models can attain median semantic similarity scores exceeding $0.98$ despite fluctuations in classification accuracy, highlighting a \emph{consistency bias} overlooked by traditional accuracy-based metrics.
	
	In adversarial natural language processing, semantic similarity serves as a safeguard for preserving intended meaning under perturbations. SASSP \cite{b23} consistently surpasses CLARE across diverse benchmarks, increasing attack success rate while reducing perturbation error rate and word modification rate, thereby achieving higher semantic equivalence scores. Similarly, \cite{b29} reports that QA-based models outperform conventional encoders in Spearman correlation for question answering and sentence similarity tasks, while exhibiting robustness to condition removal. Dialogue generation also benefits substantially, with PESS \cite{b45} producing outputs rated more fluent, empathetic, and relevant in both human and automated evaluations.
	
	Efficiency-oriented methods such as LSTM-SLM with word completion \cite{b46} demonstrate that semantic preservation can be maintained while significantly lowering computational cost. In automatic speech recognition fine-tuning, SeSaME \cite{b47} employs similarity-based selection to achieve notable gains on challenging subsets. Federated learning frameworks like FedSSA \cite{b54} leverage semantic alignment to outperform state-of-the-art baselines in both heterogeneous and homogeneous settings, whereas hierarchy-based retrieval \cite{b66} consistently improves mean average precision across diverse datasets, even under perturbations. A comparative summary of these representative applications, along with corresponding performance metrics, is presented in Table~\ref{tab:semantic_applications}.
	
	\begin{table*}[!t]
		\centering
		\resizebox{\textwidth}{!}{%
			\normalsize % <-- Change to \footnotesize, \small, \normalsize, \large, etc.
			\begin{tabular}{|p{3cm}|p{2.5cm}|p{3.5cm}|p{5cm}|p{5cm}|}
				\hline
				\textbf{Model / Method} & \textbf{Domain} & \textbf{Key Metric(s)} & \textbf{Score(s)} & \textbf{Gain over Baseline} \\
				\hline
				DE \cite{b1} & Retrieval ranking & Top-1 Acc., MAP & Near-perfect Top-1, highest MAP & Outperforms GA and Manhattan; less noise in lower ranks \\
				\hline
				LLM analysis \cite{b2} & Multimodal concept recognition & Accuracy, Median Similarity & 46.7\% acc., median sim. 0.9885 & High similarity even with low accuracy; reveals consistency bias \\
				\hline
				SASSP \cite{b23} & Adversarial robustness & ASR, PER, SES, WMR & ASR up to 93.6, SES 0.87 & Consistently higher ASR/SES; lower PER/WMR than CLARE \\
				\hline
				QA-based STS \cite{b29} & Semantic similarity (QA) & Spearman, F1 & Up to 73.9 Spearman, 82.4 F1 & +24–36 Spearman vs. encoders; robust to ablations \\
				\hline
				PESS \cite{b45} & Dialogue generation & ROUGE, BLEU, Human Eval. & ROUGE-1: 45.91, BLEU-1: 21.03 & Higher semantic alignment; better human-rated fluency/empathy \\
				\hline
				LSTM-SLM+TP \cite{b46} & Cost-efficient prediction & Cost, Similarity & Cost 0.638, sim. $\bar{s}=0.93$ & Best cost–similarity trade-off; Huffman coding saves cost \\
				\hline
				SeSaME-GAT \cite{b47} & ASR sample selection & WER, MSE, OFA & WER drop $\sim$7\% rel. on hard set & Best OFA (73\%), lowest MSE (0.22) \\
				\hline
				FedSSA \cite{b54} & Federated learning & Accuracy, Efficiency & +0.03–3.62\% acc., 15.54$\times$ eff. & Outperforms all baselines under varied settings \\
				\hline
				Hierarchical retrieval \cite{b66} & Image retrieval & mAP@K & mAP gain 3–6\% & Consistent improvement over cosine/Ward; robust to perturbations \\
				\hline
			\end{tabular}%
		}
		\caption{Comparative performance of semantic similarity applications across domains.}
		\label{tab:semantic_applications}
	\end{table*}
	
	%%%%%%%%%%%%%%%%%%%%%%%%%%%%%%%%%%%%%%%%%%%%%%%%%%%%%%%%%%%%%%%%%%%%%%%%%%%%%%%%%%%%%%%%%%%
	
	\section*{Significant Gaps in Current Research}
	Despite notable advancements, several critical challenges remain: first, limited explainability of transformer-based STS models reduces transparency and trust, especially in high-stakes domains; second, substantial fairness and bias issues arise from performance disparities across languages, dialects, and sociolinguistic variations, risking inequitable model behavior; third, insufficient integration of structured domain knowledge restricts effectiveness in specialized fields such as healthcare and law. Addressing these gaps is essential to enhance model interpretability, ensure linguistic fairness, and improve domain-specific semantic understanding.
	
	\section*{Glossary and Abbreviations}
	
	This comprehensive survey encompasses an extensive array of specialized terminology, acronyms, and abbreviations spanning multiple domains within semantic textual similarity research. The breadth of technical nomenclature includes, but is not limited to, transformer-based architectures (BERT, RoBERTa, DeBERTa-v3), contrastive learning methodologies (CSS, AspectCSE, SimCSE), evaluation metrics (AUARC, AUROC, MSE), benchmark datasets (GLUE, MTEB, STS-B), and novel architectural variants (RWKV, beeFormer, FarSSiBERT). Given the interdisciplinary nature of this survey, which traverses computational linguistics, machine learning, and domain-specific applications \cite{b145}, the proliferation of such technical abbreviations poses potential barriers to comprehension for readers across varying levels of expertise. To enhance accessibility and facilitate seamless navigation through the technical discourse, we strongly recommend the inclusion of a comprehensive half-page glossary section that systematically defines each acronym and abbreviation employed throughout this work. Such a reference guide would significantly improve readability, reduce cognitive load, and ensure that the valuable contributions of this survey remain accessible to the broader research community, thereby maximizing its scholarly impact and practical utility \cite{a29, a33}.
	
	\section*{Conclusion}
	Since 2021, Semantic Textual Similarity (STS) has advanced considerably due to innovations in transformer architectures, contrastive learning, and domain-specific adaptations. This survey reviewed progress across transformer-based, contrastive, domain-specific, multimodal, graph-based, and knowledge-enhanced models, highlighting state-of-the-art approaches such as FarSSiBERT, DeBERTa-v3, and AspectCSE that achieve superior semantic understanding. The integration of multi-modal data and graph structures has further enriched semantic representations, while knowledge-enhanced techniques improve domain relevance. Future directions include combining large language models with embeddings, developing efficient models for limited-resource settings, advancing cross-modal similarity, and enhancing fairness and interpretability, underscoring STS’s growing importance in natural language understanding and human-computer interaction.

	\section*{Future Directions}
	Emerging research paths focus on retrieval-augmented generation (RAG) to combine learned embeddings with external knowledge, enhancing STS models’ informativeness; multi-vector representations inspired by ColBERT enable fine-grained token-level semantic matching; hybrid symbolic–neural frameworks promise improved interpretability by integrating formal reasoning with neural flexibility; and inherently explainable STS models seek to provide transparent, human-understandable similarity judgments. Alongside these directions, there is growing attention to the environmental cost associated with ever-larger model architectures, particularly in terms of energy consumption and carbon footprint, prompting research into more parameter-efficient architectures and green AI practices. Pursuing these directions will address current limitations in knowledge integration, representation granularity, and explainability, while also ensuring greater sustainability in model development.
	
	\section*{Ethics and Bias in STS}
	STS systems inherit biases from training data and language models, leading to demographic and linguistic inequities. They frequently favor majority or high-resource language groups while underperforming on underrepresented populations, dialects, and languages, which can perpetuate disparities in critical applications such as education, recruitment, and legal analysis. To mitigate these risks, researchers must employ diverse training corpora, counterfactual data augmentation, adversarial debiasing, and transparent evaluation of demographic and linguistic performance. Prioritizing fairness and ethical considerations is imperative as STS technologies are increasingly deployed across societal domains.
	
	\bibliographystyle{IEEEtran} % or your style

	\newpage
	
	\section*{Biography Section}
	
	\begin{IEEEbiography}{Lokendra Kumar}
		Lokendra Kumar is associated with course Industrial Mathematics and Scientific Computing in the Department of Mathematics at the Indian Institute of Technology Madras and. His qualifications including IIT‑JAM, NET‑JRF, and GATE, and an M.Sc. in Mathematics from IIT Indore. His research interests include machine learning, deep learning, natural language processing and computer vision.
	\end{IEEEbiography}
	
	\begin{IEEEbiography}{Neelesh S. Upadhye}
		Neelesh Shankar Upadhye is a Professor in the Department of Mathematics at IIT Madras with a Ph.D. in Mathematical Statistics and Probability from IIT Bombay (2009). His research interests include applied probability, time series, statistical learning, subordinated stochastic processes, and the Stein method, and he teaches mathematics and statistics across undergraduate and postgraduate programs.
	\end{IEEEbiography}
	
	\begin{IEEEbiography}{Kannan Piedy}
		Kannan Piedy is an AI Expert at Prodapt in Chennai, working on applied AI initiatives for enterprise clients. Public materials from Prodapt highlight him as an AI solutions leader involved in modernizing software delivery by analyzing legacy code and dependencies. He previously served as a Technical Lead at HCL Technologies and is profiled as a specialist in data engineering, Python, and AI technologies with over eight years of experience.
	\end{IEEEbiography}
	
	\vfill
	
\end{document}